\documentclass[10pt,twocolumn,letterpaper]{article}

\usepackage{pifont}
\usepackage{colortbl} 
\usepackage{booktabs}
\usepackage{multirow}
\usepackage{array}
\usepackage{booktabs}
\usepackage{dsfont}
\usepackage{bm}
\usepackage{amsmath}
\usepackage{amssymb}

\usepackage{pifont}% http://ctan.org/pkg/pifont
\newcommand{\xmark}{\ding{55}}%
\definecolor{Gray}{gray}{0.9}
\definecolor{Gray2}{gray}{0.7}

\newcommand{\ie}{{\it i.e.}}
\newcommand{\eg}{{\it e.g.}}

\usepackage{colortbl} 
\usepackage{amsfonts}       % blackboard math symbols
\usepackage{nicefrac}       % compact symbols for 1/2, etc.
\usepackage{microtype}      % microtypography
\usepackage{graphics}
\usepackage{mathtools}
\usepackage{caption}
\usepackage{subcaption}
\usepackage{wrapfig,lipsum,booktabs}

\usepackage{cvpr}              % To produce the CAMERA-READY version
\usepackage[dvipsnames]{xcolor}
\definecolor{cvprblue}{rgb}{0.21,0.49,0.74}
\usepackage[pagebackref,breaklinks,colorlinks,citecolor=cvprblue]{hyperref}

\def\paperID{11211} % *** Enter the Paper ID here
\def\confName{CVPR}
\def\confYear{2024}

\title{Image-to-Image Matching via Foundation Models: A New Perspective for  Open-Vocabulary Semantic Segmentation}

\author{
Yuan Wang$^{1}$$^{*}$\qquad 
Rui Sun$^{1}$\thanks{Equal contribution}~ \qquad
Naisong Luo$^{1}$\qquad
Yuwen Pan$^{1}$\qquad
Tianzhu Zhang$^{1,2}$\thanks{Corresponding author}~ \qquad \\
$^{1}$University of Science and Technology of China\\
$^{2}$Deep Space Exploration Laboratory\\
 {\tt\small \{wy2016, issunrui, lns6, panyw\}@mail.ustc.edu.cn, \{tzzhang\}@ustc.edu.cn}  }
\begin{document}

\maketitle
\begin{abstract}
Open-vocabulary semantic segmentation (OVS)  aims to segment images of arbitrary categories specified by class labels or captions. However, most previous best-performing methods, whether pixel grouping methods or region recognition methods, suffer from false matches between image features and category labels. We attribute this to the natural gap between the textual features and visual features.
In this work, we rethink how to mitigate false matches from the perspective of image-to-image matching and propose a novel relation-aware intra-modal matching (RIM) framework for OVS based on visual foundation models. RIM achieves robust region classification by firstly constructing diverse image-modal reference features and then matching them with region features based on relation-aware ranking distribution.
The proposed RIM enjoys several merits. First, the intra-modal reference features are better aligned, circumventing potential ambiguities that may arise in cross-modal matching.
Second, the ranking-based matching process harnesses the structure information implicit in the inter-class relationships, making it more robust than comparing individually.
Extensive experiments on three benchmarks demonstrate that RIM outperforms previous state-of-the-art methods by large margins, obtaining a lead of more than 10\% in mIoU on PASCAL VOC benchmark.
\end{abstract}

\section{Introduction}
Aiming at allocating semantic labels to the corresponding pixels, semantic segmentation has achieved conspicuous achievements attributed to the development of large-scale datasets~\cite{coco,cordts2016cityscapes,ade,kirillov2023segment} and elaborate algorithms~\cite{fcn,deeplabv2,mask2former}.
However, the capabilities of conventional semantic segmentation models are restricted to predefined training categories, failing to recognize a broader spectrum of concepts, which poses a severe limitation for their practical applications.
In pursuit of the human-like intelligence of unbounded and fine-grained scene understanding, open-vocabulary segmentation (OVS)~\cite{bucher2019zero,xu2022simple} has attracted increasing interest recently, which enables segmentation with an unrestricted vocabulary.

\label{sec:intro}
\begin{figure}[t!]
	%\centering
	\includegraphics[width=\linewidth]{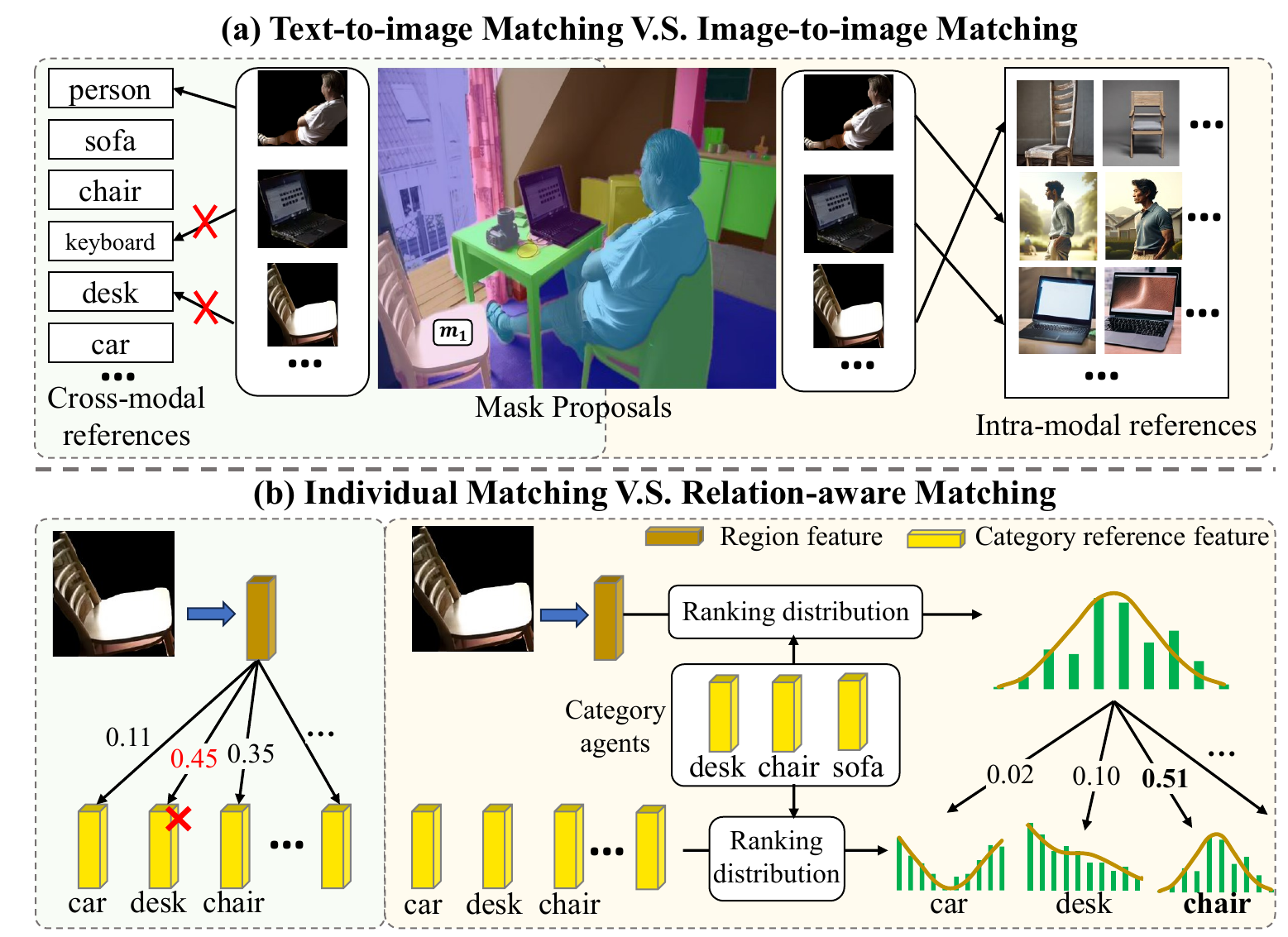}
	%\centering
	\vspace{-7mm}
	\caption{
		Motivation of our method. (a) False matches tend to occur in cross-modal features. We establish well-aligned image-modal reference features thus transit the text-to-image matching to image-to-image matching. (b) Indivdual matching 
  tends to suffer from disturbances. We propose a novel relation-aware matching strategy for more robust region classification.
}\label{fig:motivation}
	\vspace{-7mm}
\end{figure}

OVS is highly challenging as it requires not only grouping all pixels to corresponding visual concepts but also assigning each region the correct semantic label from a large vocabulary.
Some early attempts~\cite{xu2022groupvit, luo2023segclip} optimize the pixel grouping process and the classification of visual concepts simultaneously by employing contrastive learning with image-text pairs.
These methods inevitably yield masks of relatively low quality with only coarse supervision available.
Another line of works~\cite{xu2022simple, san, mask-adapted-clip} significantly dominate this field by modeling the OVS as a region recognition problem, which decouple OVS into two procedures, \ie class-agnostic mask proposals prediction and mask class recognition. Though the mask proposals generation has been significantly improved with the assistance of pixel-level supervision and elaborate segmenter architectures (\eg Mask2Former~\cite{mask2former}), the unsatisfactory region class recognition has emerged as the performance bottleneck of OVS~\cite{mask-adapted-clip}.
We attribute this to the natural gap between the highly abstract and monotonous category textual features and the visual features that are more concrete and diverse.

Most current OVS approaches achieve mask class recognition via cross-modal region-category alignment based on the pretrained vision-language models (\eg, CLIP~\cite{clip}).
Despite its exemplary generalization ability on downstream classification tasks, CLIP usually suffer from spatial relations ambiguity~\cite{oside,yi2023simple} and co-occurring object confusion ascribed to the holistic pre-training objective.
On the other side, also pre-trained on internet-scale data, Stable Diffusion~\cite{Stable_diffusion} (SD) model has garnered growing research interest credited to the phenomenal power of synthesizing photorealistic images with diverse and plausible content conditioned on textual descriptions. %as well as the strong correspondence between the visual and textual features embedded within the model.
Considering that features from the same modality inherently exhibit a higher degree of alignment compared to cross-modal features (as shown in Fig.~\ref{fig:tsne}), we ask the question: \ding{182} \textit{is it possible to transition the region classification in OVS from image-to-text matching to image-to-image matching using the SD model?}

Before tackling this issue, we first revisit the common practices in mask class recognition~\cite{xu2022simple,mask-adapted-clip}, which involves matching visual concepts with a set of category reference features (\eg category textual features from CLIP). 
Inspired by previous exploration~\cite{wu2023diffumask,wu2023datasetdm,zhao2023unleashing} of the strong image-text correspondence exhibited by the SD model, we can establish better-aligned intra-modal category reference features as shown in Fig.~\ref{fig:motivation} (a).
The cross-attention maps in the conditional generation process could serve as exceptional tools to further refine the category features.
Nonetheless, it is still non-trivial to map test region features to their corresponding category reference features precisely, as the intra-class diversity and disturbances from akin categories exhibit in practical scenarios potentially lead to mismatches.
Previous methodologies process each category independently during the matching phase, overlooking the informative inter-class relationships, which implicitly integrate structured contextual information and effectively aid in disambiguation. Here another question arise: \ding{183} \textit{how to harness the structure information modeled in the inter-class relationship to facilitate more accurate matching?}

Driven by this question, we carefully design the relation-aware similarity measurement that incorporates the relations of the current region with some semantic relevant category agents. % into the matching procedure.
For a specified region feature within test image, a subset of category reference features that most akin to the region are selected to serve as category agents.
The central idea is that we regard the category agent ranking as a stochastic event rather than a deterministic permutation.
For example, in ranking, given the region feature specified by the mask proposal $m_1$ in Fig.~\ref{fig:motivation} (a), its scores vary for different category agents, which can be taken as probabilities.
The probability of being ranked first is 0.45 of the agent `desk' and 0.35 of the agent `chair'. The ranking permutation reflects the relevance of the corresponding categories w.r.t. the region feature. 
An agent-ranking probability distribution can be constructed by associating the probability with every rank permutation for both the region feature and all category reference features.
Finally, as illustrated in Fig.~\ref{fig:motivation} (b), we transform the similarity measurement from individual region-reference comparison to relation-aware agent-ranking distribution similarity.

% In this paper, we present a vision foundation model-based OVS framework DiMO (Stable-\textbf{Di}ffusion~\cite{Stable_diffusion} \& SA\textbf{M}~\cite{kirillov2023segment} \& DIN\textbf{O}v2~\cite{oquab2023dinov2}) that coherently addresses the questions \ding{182}-\ding{183}.
%
In this paper, we present a training-free OVS framework RIM to achieve \textbf{R}elation-aware \textbf{I}ntra-modal \textbf{M}atching based on visual foundation models that coherently addresses the questions \ding{182}-\ding{183}.
Specifically, we facilitate a synergistic collaboration between the SD model and SAM to generate category-specific reference features by prompting the Segment Anything Model~\cite{kirillov2023segment} (SAM) with points selected from the cross-attention map within SD model. SAM is further adopted to provide mask proposals of the test images. Finally, we conduct the relation-aware matching based on ranking distribution in the robust all-purpose feature space of DINOv2~\cite{oquab2023dinov2}.
RIM ensembles expertise of different visual foundational models in a complementary manner,  enhancing mask quality and region-category matching precision simultaneously. Moreover, the overall framework is training-free, substantially mitigating the risk of overfitting.

Our contributions can be concluded as follows: (1) We reveal the problems of region feature classification in OVS and propose to construct well-aligned intra-modal reference features to circumvent ambiguities of cross-modal matching.
(2) We design a relation-aware matching strategy based on ranking distribution, which captures structure information implicit in inter-class relationships and enables more robust matching.
(3) We propose a training-free relation-aware intra-modal matching (RIM) network for OVS based on visual foundation models. Extensive experiments demonstrate that RIM remarkably surpasses previous state-of-the-art methods by a large margin.

\section{Related Work}
\label{sec:rw}
\textbf{Semantic Segmentation} is a fundamental computer vision task
with widespread applications in fields such as medical image processing~\cite{wangkai2023maunet, sun2021lesion,luo2024electron, sun2023structure,pan2023adaptive,ijcai2023p158}, video analysis~\cite{sun2023alignment,sun20221st}.
The pioneering FCN~\cite{fcn} has inspired a multitude of subsequent endeavors~\cite{unet,pspnet,xiao2018unified,deeplab,deeplabv2} centered around Convolutional Neural Networks (CNN).
%
%Numerous studies have devised intricate architectures that focus on better context exploration by enlarging the receptive field  of CNN via global pooling~\cite{parsenet}, pyramid pooling~\cite{yang2018denseaspp,deeplabv2}, or dilated convolutions~\cite{deeplabv2,deeplab}.
%
Beyond the CNN-based models, the triumph of the Vision Transformer (ViT) has spurred a succession of transformer-based segmentation models~\cite{segformer,segmenter,early,luo2023camouflaged}, which have progressively evolved into a unified segmentation framework~\cite{maskformer, mask2former} capable of addressing various segmentation tasks.
Some alternative settings, such as few-shot~\cite{asgnet,wang2022adaptive,wang2023focus,wang2023rethinking}, semi-supervised~\cite{asgnet,sun2023daw,mai2023dualrel,mai2024pay}, or weakly supervised~\cite{huang2018weakly,wang2020self} segmentation, attempt to enhance the practicality of semantic segmentation techniques. Despite their success, these models are confined to predefined training categories or the specific foreground class, failing to recognize a broader spectrum of categories. 

\textbf{Open-vocabulary Semantic Segmentation} aims to segment images with arbitrary categories described by texts.
Some efforts~\cite{bucher2019zero, xian2019semantic, yi2023simple,luo2023segclip,ovsegmenter} concentrated on developing a joint embedding space that bridges image pixels with class names or descriptions via learning objectives established from image-text pairs. 
For example, GroupViT~\cite{xu2022groupvit} designed a hierarchical grouping transformer and learned the alignment between groups and text via contrastive loss.
Another line of works~\cite{mask-adapted-clip, xu2022simple, oside} models OVS as a region recognition task by decoupling it into mask proposal generation and region classification,  achieving notable progress attributed to advanced segmentation architectures~\cite{mask2former} and pixel-level annotations~\cite{coco,ade}.
Large-scale vision-language pre-training models such as CLIP~\cite{clip} and ALIGN~\cite{align} have been widely applied for OVS~\cite{fc-clip, san, one-pass} which endows OVS model enhanced generalization. However, those pre-training models often encounter spatial confusion in dense prediction tasks~\cite{yi2023simple}, leading to recurring misclassification issues, which has emerged as a critical bottleneck in OVS.

\textbf{Visual foundation models} is catching up with the research in natural language processing (NLP)~\cite{brown2020language,chowdhery2022palm,ouyang2022training} and have achieved conspicuous
achievements across a wide range of visual tasks. For example, DINOv2~\cite{oquab2023dinov2} establishes an impressive all-purpose feature extractor via self-supervised learning at both the image and patch levels. Trained on over 1 billion masks, Segment Anything Model (SAM)~\cite{kirillov2023segment} has demonstrated astonishing zero-shot class-agnostic segmentation performance. Diffusion models~\cite{ho2020denoising,song2020score} have propelled the advancement of a series of image-to-text generation systems such as DALL-E~\cite{dalle} and Stable Diffusion model~\cite{Stable_diffusion}.
A series of works~\cite{liu2023matcher,oside,wu2023diffumask} render visual foundation models as powerful out-of-the-box tools to handle downstream tasks. 
%For instance, diffuMask~\cite{wu2023diffumask} exploits the stable diffusion model to synthesize segmentation datasets and achieves a competitive performance over the counterpart of real data. 
Though a single foundation model may have limited capacity in addressing complex visual tasks such as OVS, in this work, we demonstrate that integrating different foundation models leads to positive synergies.
\section{Method}
\label{sec:Method}
\begin{figure*}[t!]
	\centering
	\includegraphics[width=0.98\linewidth]{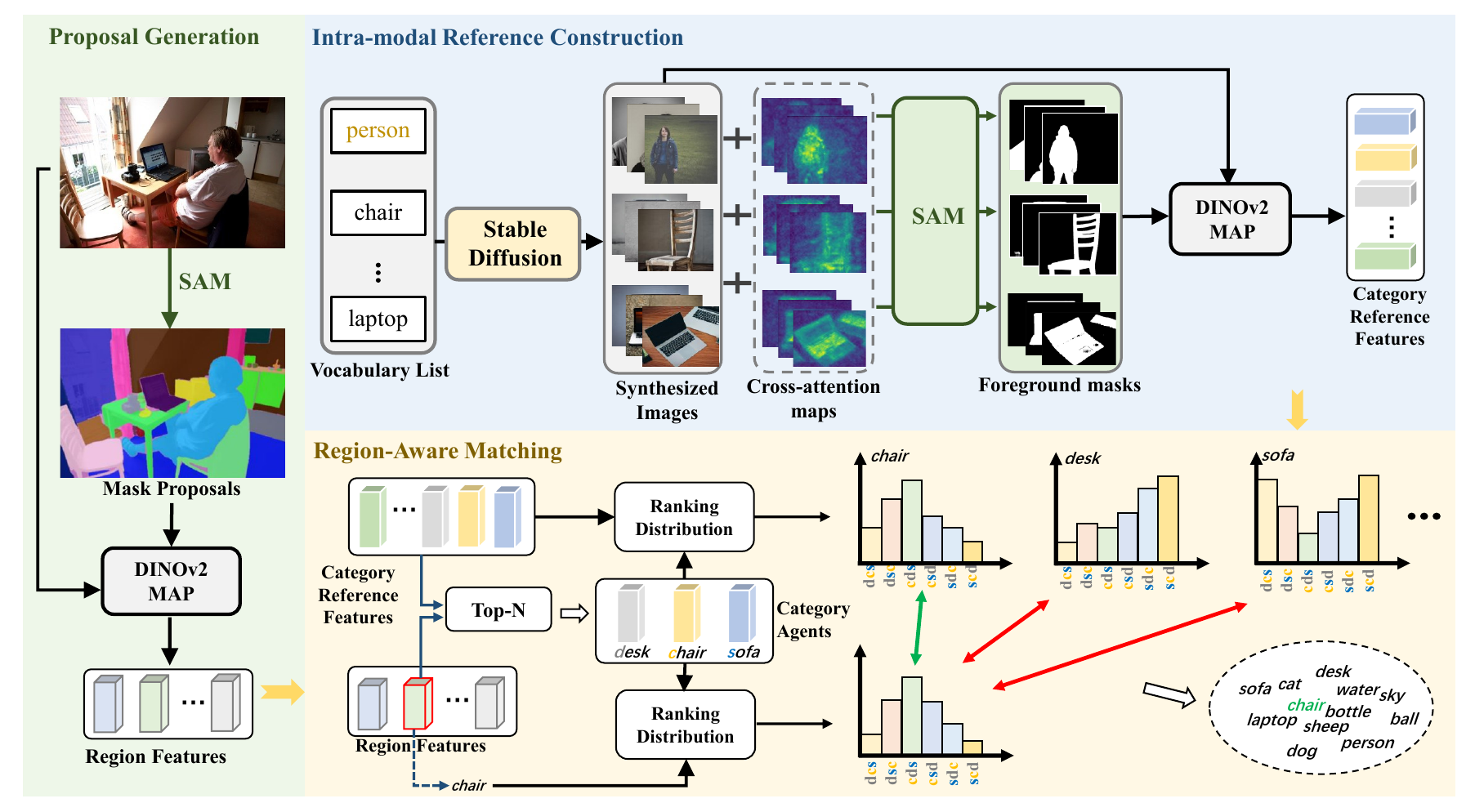}
	%\centering
	\vspace{-3mm}
	\caption{
		Framework of our proposed Relation-aware Intra-modal Matching (RIM) Network. We first explore Stable Diffusion model and SAM to construct image-modal reference features, then we conduct relation-aware matching between region features and reference features based on ranking distribution. The matching is established in the all-purpose feature spaces of DINOv2.
	}
	\vspace{-5mm}
 \label{fig:framework}
\end{figure*}
\subsection{Problem Definition}
%OVsegmenter主要用于给定一张图像和一个类别集合，将所有的类别分割出来。之前的方法主要在训练集C_train中进行训练,在测试集C_test上进行测试，其中C_train ≠ C_test.通常在C_train中，包含着不同程度的标注信息，例如图像的类别标签或者文本描述，甚至类别绑定的mask标签。在本文中，我们探索一个更加一般化的zero-shot training-free的方式。所有的测试类别均是novel类别。
Open-vocabulary segmentation aims to segment any image against a new vocabulary of categories $\mathbf{C}_{test}$. Previous approaches typically involve training models on datasets annotated with image-level textual labels or pixel-level masks, with the category set $\mathbf{C}_{train}$. $\mathbf{C}_{test}$ contains novel categories not exposed to the training process, \ie, $\mathbf{C}_{train} \neq \mathbf{C}_{test}$.
Owing to the training-free nature of our method, it inherently operates under a more challenging zero-shot setting wherein all test categories are considered novel.

\subsection{Preliminary of Vision Foundation Models}
\textbf{Stable-Diffusion model.} Different from discriminative image-text models that model the class probability distribution $\boldsymbol{p}(\boldsymbol{c}|\boldsymbol{i})$ given the image, text-to-image stable diffusion model encode a text-conditional distribution of possible images $\boldsymbol{p}(\boldsymbol{i}|\boldsymbol{c})$.  This model demonstrates the prowess to synthesize high-fidelity images $\mathbf{I} \in \mathbb{R}^{H \times W \times 3}$  that closely align with the specified conditional text $\mathcal{T}$, all originating from a latent space defined by random Gaussian noise $\mathbf{z} \sim \mathcal{N}(0,1)$.
Specifically, stable diffusion model consists of three components: a pre-trained variational autoencoder (VAE)~\cite{vae} that encodes and decodes the image latent code, a text encoder for prompt embedding, and a time-conditional UNet~\cite{unet} for the denoising of latent vectors.
The visual-text interaction occurs in the cross-attention layers that are integral to the UNet structure for each denoising step.
Target categories within synthesized images can be accurately localized based on class-discriminative cross-attention maps.

\textbf{Segment Anything Model.} SAM comprises of three components: an image encoder $\mathbf{Enc_{I}}$, a prompt encoder $\mathbf{Enc_{P}}$, and fast mask decoder $\mathbf{Dec_{M}}$. 
$\mathbf{Enc_{P}}$ takes as input the prompts of various optional forms such as points, boxes or coarse masks, and translates the prompts into feature tokens $\mathbf{F_{P}}$.
Image feature tokens $\mathbf{F_{I}}$ are also obtained from $\mathbf{Enc_{I}}$.
A series of learnable mask tokens $\mathbf{F_{M}}$ are introduced and concatenated with $\mathbf{F_{P}}$ for diverse masks generation.
Then the light weight $\mathbf{Dec_{M}}$ integrates the $\mathbf{F_{I}}$ and the concatenation of $\mathbf{F_{P}}$ and $\mathbf{F_{M}}$ to predict segmentation masks.

\textbf{DINOv2.} Pretrained on a large quantity of curated data with image and patch level distriminative self-supervised learning, DINOv2~\cite{oquab2023dinov2} learns all-purpose visual features that work out of box on many tasks varing from image level, \eg, image classification and pixel level, \eg, semantic segmentation. Moreover, this all-purpose ViT model demonstrates impressive patch-matching ability, robustly capturing similar semantic intent across different objects and even distinct images~\cite{liu2023matcher}.

\subsection{Overview}
As shown in Fig.~\ref{fig:framework}, we propose a training-free OVS framework RIM based on visual foundation models. 
RIM tackles the challenging region classification in OVS from a novel image-to-image matching perspective via the following two procedures, \ie, 1) intra-modal reference features construction, and 2) relation-aware matching. We resort to the SD model~\cite{Stable_diffusion} and the SAM~\cite{kirillov2023segment} to establish category reference features in procedure 1).
In procedure 2), we conduct ranking based matching between the reference features and the region features in the all-purpose DINOv2 feature space.
% Here the region features are specified by mask proposals, which are obtained from SAM by sampling point prompts in a grid over the image.
%
The details are as follows.

\subsection{Intra-modal Reference Construction}
In contrast to the substantial gap between abstract, monotonous text features and the concrete, diverse image features, homogenous modality features naturally exhibit improved alignment characteristics as shown in Fig.~\ref{fig:tsne}.
We resort to stable diffusion model to generate a handful of reference images $\mathcal{I}=\{\mathbf{I}^c_1, \mathbf{I}^c_2, ..., \mathbf{I}^c_K | \mathbf{I}^c_k \in \mathbb{R}^{H\times W \times 3}, c=1,2,\dots, C\}$ for all $C$ candidate classes by simply prompting it with ``a photo of [category name]''.
Together with the images, we also acquire the corresponding cross-attention maps for localizing the foreground targets.
Specifically, for time step $t$, the noisy image features $\mathbf{F}_v \in \mathbb{R}^{h\times w \times c}$ are flattened and linearly projected into the $queries$ $\mathbf{Q}$ and the prompt features $\mathbf{F}_p \in \mathbb{R}^{N\times d}$ are respectively projected into the $keys$ $\mathbf{K}$ and $values$ $\mathbf{V}$:
\begin{equation}
    \mathbf{Q} = \mathbf{F}_v\mathbf{W}^{\mathcal{Q}},\quad
    \mathbf{K} = \mathbf{F}_p\mathbf{W}^{\mathcal{K}},\quad
    \mathbf{V} = \mathbf{F}_p\mathbf{W}^{\mathcal{V}}.
\end{equation}
Among which, $\mathbf{W}^{\mathcal{Q}}$, $\mathbf{W}^{\mathcal{K}}$, $\mathbf{W}^{\mathcal{V}}$ are linear projections. The cross-attention maps are calculated as:
\begin{equation}
    \mathcal{S} = \mathbf{Softmax}(\frac{\mathbf{Q}\mathbf{K}^\mathsf{T}}{\sqrt{d}}),
        \label{softmax}
\end{equation}
where the $\sqrt{d}$ is the scaling factor.
Cross-attention maps $\mathcal{S}_{n}^{l,t}$ of different text tokens from different layers of UNet can be obtained based on the Equ.~\ref{softmax} and $n$,$l$,$t$ are the index of text tokens, UNet layers and diffusion steps, respectively.
To obtain the cross-attention map corresponding to the category token for foreground mining, we follow~\cite{wu2023diffumask} to aggregate normalized multi-time and multi-layer maps, formaly,
\begin{equation}
    \Bar{\mathcal{S}}_n =  \frac{1}{L\cdot T}\sum_{l\in L, t\in T} \frac{\mathcal{S}_{n}^{l,t}}{\mathbf{max}(\mathcal{S}_{n}^{l,t})},
\end{equation}
Where the $L$ and $T$ denote the total time steps and the number of UNet layers.

The foreground area in the synthesized images should be further located to avoid the interference of the irrelevant background regions. However, simply binarizing the cross-attention maps by thresholding may fail to fully capture the target. We thus exploit SAM to generate target masks by sampling the prompt points within the binarized attention maps.
The foreground masks $\mathcal{M}=\{\mathbf{M}^c_1, \mathbf{M}^c_2, ..., \mathbf{M}^c_K | \mathbf{M}^c_k \in \mathbb{R}^{h\times w}, c=1,2,\dots, C\}$ and corresponding images $\mathcal{I}$ then serve as the class references.% better-aligned  of relation-aware matching.

To enhance the robustness of instance-level matching, we opt to construct class reference image features within the all-purpose feature space of DINOv2~\cite{oquab2023dinov2} via mask average pooling (MAP). Specifically,
\begin{equation}
    \mathcal{F}_{ref} =\{ \mathbf{F}^c|\mathbf{F}^c=\frac{1}{K}\sum_{k\in K}\mathbf{MAP}(\varphi(\mathbf{I}_k^c), \zeta(\mathbf{M}_k^c))\}_{c=1}^C
\end{equation}
where the $\varphi$ and $\zeta$ denote the DINOv2 feature extractor (ViT) and the bilinear-interpolation resize function, respectively. The $\mathbf{F}^c \in \mathbb{R}^{1 \times D}$ represents the average of all synthesized foreground features. We also construct a background reference feature by averaging all the background features of synthesized images. Similarly, region features within the test image specified by mask proposals are obtained as: 
\begin{equation}
    \mathcal{F}_{test} = \{\mathbf{F}^p|\mathbf{F}^p=\mathbf{MAP}(\varphi[\mathbf{I}_{test}], \zeta[\mathbf{M}^p])\}_{p=1}^P,
\end{equation}
among which $\mathbf P$ is the number of masks generated by SAM.
By designing collaborative interactions among various visual foundational models, we establish well refined image-modal reference features thus shifts the paradigm of region classification in OVS from cross-modal matching to better-aligned intra-modal matching.

\subsection{Relation-aware Matching}
\label{sec: matching}
The naive region classification strategy only selectively recruits the class reference feature with the highest similarity. 
However, independently taking each category exacerbates the risk of mismatches owing to the intra-class diversity and interference from similar categories.
%
%We argue that inter-class relationship should be further considered for more effective matching.
%
To further harness the structure information implicit in inter-class relationship for more effective matching, we select the top N reference features most similar to the region feature $\mathbf{F}^p$ as category agents $\mathbf{A}^p \in \mathbb{R}^{N \times D}$.
We derive the scores of region-agent relation $s^p \in \mathbb{R}^{1\times N}$ between the region feature $\mathbf{F}^p$ and the category agents $\mathbf{A}^p$ using cosine similarity:
\begin{equation}
    \mathbf{s}^p = \frac{\mathbf{F}^p(\mathbf{A}^p)^\mathsf{T}}{\Vert \mathbf{F}^p\Vert_2 \cdot \Vert \mathbf{A}^{p} \Vert_2}.
    \label{eq:6}
\end{equation}
The core idea is that we take the agent ranking as a random event rather than a deterministic permutation. This implies that each permutation of the category agents is present with a certain probability, as opposed to the exclusive existence of the permutation ordered from largest to smallest.
The probability of one permutation $\pi \in \mathcal{P}(|\mathcal{P}|=N!)$ given $\mathbf{s}$ (we omit region index $p$ for brevity) can be calculated as:
\begin{equation}
   P(\pi \vert \mathbf{s}) = \prod _{k=1}^K \frac{\mathbf{s}_{\pi(k)}}{\sum _{k'=k}^K\mathbf{s}_{\pi(k')}}, 
  \label{eq:7}
\end{equation}
among which the $\pi(k)$ denotes the $k^{th}$ class index of this permutation.
For instance, assume that for a given region, the selected agents correspond to the categories ``sofa'', ``chair'', and ``bench'' respectively. One of the permutations of these three agents is $\pi=(chair,sofa,bench)$. We can derive the probability of $\pi$ based on the region-agent relation $\mathbf{s}$:
\begin{equation}
   P(\pi \vert \mathbf{s}) = \frac{\mathbf{s}(\textit{chair})}{\mathbf{s}(\textit{sofa})+\mathbf{s}(\textit{bench})+\mathbf{s}(\textit{chair})}\cdot \frac{\mathbf{s}(\textit{sofa})}{\mathbf{s}(\textit{sofa})+\mathbf{s}(\textit{bench})}.  
  \label{eq:8}
\end{equation}
By associating the probabilities of all $|\mathcal{P}|$ permutations, We convert the scores of individual region-agent relation $\mathbf{s}$ into class ranking probability distributions $P(\pi \in \mathcal{P}|\mathbf{s}^p) \in \mathbb{R}^{1\times |\mathcal{P}|}$, effectively modeling the inter-class relationship.
Similarly, we compute the agent-ranking probability distributions for each category reference features on the same category agents, resulting in $P(\pi \in \mathcal{P}|\mathbf{s}^r) \in \mathbb{R}^{1\times |\mathcal{P}|}$, where $\mathbf{s}^r$ is obtained through Equ.~\ref{eq:6}. The distribution of the region is compared against the distributions of all reference features via cosine similarity, enabling the determination of the classification result:
\begin{equation}
    cls = \underset{r=1,2,\dots,C}{\arg\max}[\mathbf{cosine}(P(\pi \in \mathcal{P}|\mathbf{s}^p),
    P(\pi \in \mathcal{P}|\mathbf{s}^r))].
\end{equation}

To further leverage the advantage of diversity in image-modal reference features, we construct a series of subcategory reference features by clustering the foreground prototypes (obtained by mask average pooling) of all synthesized images.
The subcategory reference features are involved in the similarity computation instead of the holistic ones, and the final score of a category is obtained by summing up the similarities of all the corresponding subcategory reference features. The classification result is then obtained by:
\begin{equation}
    cls = \underset{r=1,2,\dots,C}{\arg\max}[\sum_{t=1}^T\mathbf{cosine}(P(\pi \in \mathcal{P}|\mathbf{s}^p),
    P(\pi \in \mathcal{P}|\mathbf{s}_t^r))].
    \label{equ:10}
\end{equation}
Based on the metric grounded in distributional similarity, instances of erroneous region classification can be effectively reduced as inter-class relationships are exploited to facilitate disambiguation.
\section{Experiments}
\label{sec:Experiments}
\begin{table*}[t]
\centering
\resizebox{0.95\textwidth}{!}{
  \begin{tabular}{l|cccc|ccc}
    \toprule
    \multirow{2}{*}{Method}& 
    \multirow{2}{*}{Arch} &
    \multirow{2}{*}{Training dataset} & 
    \multirow{2}{*}{Supervision} & 
    \multirow{2}{*}{Zero-shot transfer} & 
    \multicolumn{3}{c}{Downstream datasets} \\
    \cmidrule(r){6-8}
    & & & & & PASCAL VOC & PASCAL Context & COCO Object \\
    \midrule
    \midrule
    
    DeiT \cite{deit} & ViT & IN-1K & class label & \xmark &  53.0 & 35.9 & - \\
    % \midrule
    MoCo \cite{moco} & ViT& IN-1K & self & \xmark & 34.3 & 21.3 & - \\
    DINO \cite{dino} & ViT & IN-1K & self & \xmark & 39.1 & 20.4 & - \\
    % MoCo \cite{moco} & ViT-S & CC12M+YFCC15M & self & \xmark & 36.1 & 23.0 & - \\
    % DINO \cite{dino} & ViT-S & CC12M+YFCC15M & self & \xmark & 37.6 & 22.8 & - \\
    \midrule
    ViL-Seg \cite{vilseg} & ViT & CC12M & self+text  & $\checkmark$ & 33.6 & 15.9 & - \\
    
    % CLIPpy \cite{clippy} & ViT-B & CC12M &text & $\checkmark$ & 50.8 & - & 23.8$^\dag$ \\
    % GroupViT~\cite{xu2022groupvit} & ViT-S & CC12M+YFCC15M & text & $\checkmark$ & 51.2 & 22.3 & 20.9 \\ 
    % \midrule
    % GroupViT* \cite{gvt} & ViT-S & CC4M & text & $\checkmark$ & 19.8 & 8.8 & 9.1 \\
    % GroupViT* \cite{gvt} & ViT-B & CC4M & text & $\checkmark$ & 25.8 & 11.3 & 10.7 \\
    % GroupViT* \cite{gvt} & ViT-S & CC12M & text & $\checkmark$ & 40.2 & 18.7 & 17.7 \\
    %GroupViT* \cite{gvt} & ViT-B & CC12M & text & $\checkmark$ & todo & todo & todo \\
    %FreeSeg & ViT-B & IN-1K+CC3M & self+text & $\checkmark$ & 50.3 & - & 20.4 \\
    % \midrule 
    % OVSegmentor (ours) & ViT-S & CC4M & self+text & $\checkmark$ & 44.5 & 18.3 & 19.0 \\

    MaskCLIP~\cite{maskclip} & ViT & LAION   & $text+CLIP_{\rm T}$ &$\checkmark$ & 38.8 & 23.6 & 20.6 \\
    GroupViT~\cite{xu2022groupvit}  & ViT & CC12M & text & $\checkmark$ & 52.3 & 22.4 & - \\
    ZeroSeg~\cite{zeroseg} & ViT  & IN-1K     & $CLIP_{\rm V}$ &$\checkmark$ & 40.8  & 20.4 & 20.2 \\
    TCL~\cite{tcl} &ViT& CC3M+CC12M & text        & $\checkmark$ & 51.2 & 24.3 & 30.4 \\
    ViewCo~\cite{ren2023viewco} &ViT &CC12M+YFCC &text+self &$\checkmark$ &52.4 &23.0 &23.5 \\
    CLIPpy \cite{clippy} & ViT & HQITP-134M & text & $\checkmark$ & 52.2 & - & 32.0 \\
    SegCLIP~\cite{luo2023segclip} & ViT & CC3M+COCO & $text+CLIP_{\rm T}$ &$\checkmark$ & 52.6 & 24.7 & 26.5 \\
    OVSegmentor~\cite{ovsegmenter} & ViT & CC4M & self+text & $\checkmark$ & 53.8 & 20.4 & 25.1 \\
    SimSeg~\cite{yi2023simple} &  ViT  & CC3M+CC12M & text        & $\checkmark$ & 57.4 & 26.2 & 29.7 \\
    \midrule
    DiffSeg~\cite{diffseg} & UNet+ViT &  &  &$\checkmark$ & 60.1  & 27.5 & 37.9 \\
    OVDiff~\cite{ovdiff} & UNet & \multicolumn{2}{c}{Training-free} &$\checkmark$ & 67.1  & 30.1 & 34.8 \\ \rowcolor{Gray}
    Ours & UNet+ViT &   &  &$\checkmark$ & \textbf{77.8}  & \textbf{34.3} & \textbf{44.5} \\
    
    %Ours & ViT-B & CC12M & self+text & $\checkmark$ & - & - & - \\
    %Ours & ViT-B & CC12M+YFCC15M & self+text & $\checkmark$ & TODO & TODO & TODO \\
    \bottomrule
  \end{tabular}}
  \vspace{-3mm}
      \caption{\textbf{Comparison with existing methods.}
Models in the first three rows are finetuned on target datasets while the rest approaches perform zero-shot transfer. Bold fonts refer to the best results among the models which enable zero-shot transfer. With only image-text pairs available, our method significantly outperforms the existing approaches. More results please refer to \textbf{Supplementary Material}.}
\vspace{-5pt}

%\weidi{say the .... numbers are copied from which groupvit.}}
%\vspace{-0.2cm}
% \vspace{-0.4cm}
\label{tab:sota}
\vspace{-4mm}
\end{table*}
% In this section, we first introduce the datasets and the evaluation Metrics employed in our experiments. Subsequently, we describe the details of the model architecture and the implementation of our experiments. Finally, we conduct a performance comparison between our method and previous approaches. Ablation studies are also presented to verify the effectiveness of each component of our method.
\subsection{Dataset and Evaluation Metric}
We evaluate our model on three commonly used benchmarks, namely, PASCAL VOC 2012~\cite{everingham2010pascal}, PASCAL Context~\cite{pascal0context} and COCO Object~\cite{coco}, which have 20,59,80 foreground classes, respectively. We also consider an extra background class in all three datasets.
Training sets of datasets are not needed as the proposed method is training-free. For a fair comparison with previous approaches, we directly evaluate our method on the validation sets of these datasets, including 1449, 5105, and 5000 images, respectively. We use the mean of class-wise intersection over union (mIoU) following the common practice to measure the performance.% of our approach.

\begin{table}[t]
\centering
\resizebox{\columnwidth}{!}{
\begin{tabular}{cc|cc|c}
			\bottomrule
		\multicolumn{2}{c|}{IRC}  & \multicolumn{2}{c|}{RM}   &\multirow{2}{*}{mIOU}   \\
			\cline{1-4}
			  w/o fg-mask & w/ fg-mask  &  w/o sub &  w/ sub & \\
			\hline
			\hline
			& & & & 26.7 \\
			$\checkmark$ &    & &  & 38.7  \\
			&  $\checkmark$  & &   &  41.8  \\
			  & $\checkmark$  &  $\checkmark$ &   & 43.1 \\ \rowcolor{gray!30}
			  &$\checkmark$ & &  $\checkmark$  &\textbf{44.5}  \\
			\bottomrule
	\end{tabular}}
 \vspace{-3mm}
\caption{Ablation studies of the proposed RIM. We mainly verified the effectiveness of the image reference construction (IRC), and relation-aware matching (RM). Moreover, we also observe the effectiveness of foreground segmentation (fg-mask) of synthesized images and the subcategory reference features (sub).}
\label{tab:ablation}
\vspace{-4mm}
\end{table} 

 \vspace{-3mm}
\begin{table}[t]
\centering
\resizebox{\columnwidth}{!}{
\begin{tabular}{c|cccc}
			\bottomrule
			  Encoder & CLIP~\cite{clip} & SD~\cite{Stable_diffusion} &  MAE~\cite{mae} & DINOv2 \\
			\bottomrule
              \bottomrule
			  mIoU & 42.0& 43.1 & 39.5&\textbf{44.5}  \\
			\bottomrule
	\end{tabular}}
 \vspace{-3mm}
 \caption{Comprison of different image feature extractors.}
\label{tab:encoder}
 \vspace{-2mm}
\end{table}

\begin{table}[t]
	\vspace{0.0cm}
	\vspace{-4mm}
	\begin{minipage}[t]{0.55\hsize}	
		\centering
		\resizebox{1.0\hsize}{!}{
		\begin{tabular}{c|ccc}
			\bottomrule
			  Segmenter  & none & MaskFormer & SAM \\
			\bottomrule
			\bottomrule
			  mIoU & 42.0 & 76.7 & \textbf{77.8}\\
                \bottomrule
	\end{tabular}
  }
		\vspace{-2mm}
 \caption{Comparison of mask proposal generators. ``none'' means direct pixel classification.}
   \label{tab:sementer}
	\end{minipage}
	\hspace{0.2cm}
	\begin{minipage}[t]{0.4\hsize}	
		\centering
		\resizebox{0.98\hsize}{!}{
			\begin{tabular}{c|c}
				\bottomrule
				Selection of Agents & mIoU \\
				\bottomrule
                    \bottomrule
				Rand & 43.8 \\
				Top-$N$ & \textbf{44.5}\\
				\bottomrule
		\end{tabular}}
	\vspace{-3mm}
    \caption{Comparison of category agents selection.}
      \label{tab:agents}
	\end{minipage}	
 % \vspace{-4mm}
\end{table}

\subsection{Implementation of Visual Foundation Models}
We adopt the Stable Diffusion model v1.4~\cite{Stable_diffusion} to generate category-specific images of resolution $512 \times 512$ for category reference feature construction.
We employ DINOv2~\cite{oquab2023dinov2} with a ViT-B~\cite{vit2020} as the default image encoder for more discriminative matching. Specifically, the $keys$ sequence of the last attention layer is reshaped as the feature map.
SAM~\cite{kirillov2023segment} with ViT-B is adopted as the segmenter. We collect 32x32 prompt points in a grid manner to generate mask proposals for the test images.
To segment the foreground of synthesized reference images, We sample 5 prompt points within the binarized cross-attention map, where the binarization threshold is set to a relatively high value of 0.7 to prevent erroneous segmentation of background regions. The number of category agents is set to 4.
More implementation details are provided in \textbf{Supplementary materials}.

\begin{figure*}[t!]
	\centering
	\includegraphics[width=0.94\linewidth]{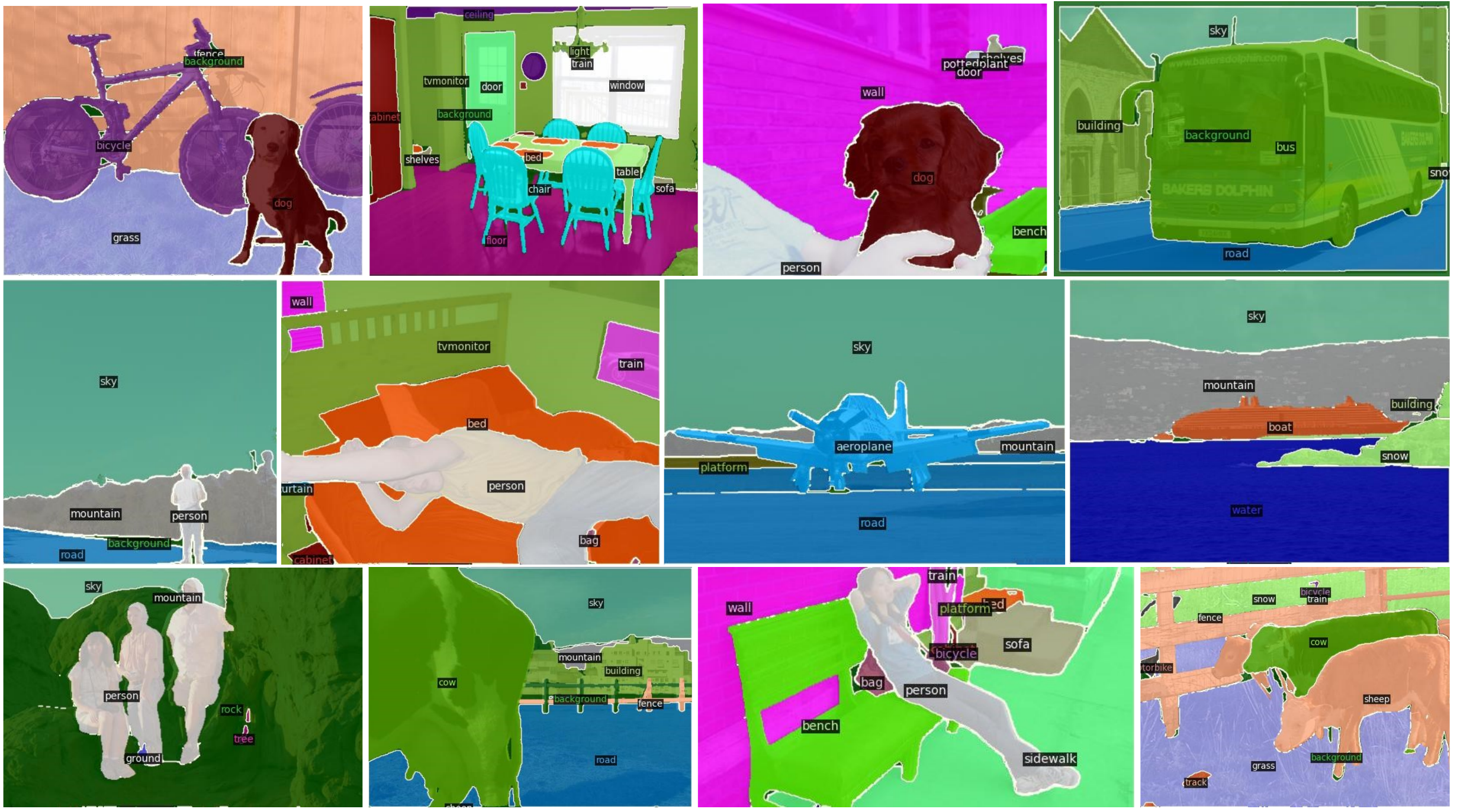}
	%\centering
	\vspace{-3mm}
	\caption{
		Qualitative results of our method. 
	}\label{overall performance}
	\vspace{-6mm}
\end{figure*}
\subsection{Comparison to the state-of-the-arts}
We compare our method with approaches that have been trained with fully supervised finetuning transfer and zero-shot transfer. Besides, we also conduct performance comparisons with newly arising train-free approaches.
Table~\ref{tab:sota} summarizes the results of the comparison. 

Firstly, we can observe that the proposed RIM significantly surpasses the non-zero-shot fully supervised segmentation baselines, \ie, DeiT~\cite{deit}, MoCo~\cite{moco}.
Furthermore, we compare with other zero-shot OVS approaches such as MaskCLIP~\cite{maskclip}, SegCLIP~\cite{luo2023segclip}, and SimSeg~\cite{yi2023simple}, which also adopt the ViT backbones of visual foundation model, \ie, CLIP~\cite{clip}. Our method also shows a clear lead. Specifically, we achieve 20.4\%, 18.1\%, 14.8\% mIoU over the SimSeg~\cite{yi2023simple} on three datasets, respectively.
We posit that the primary cause is that CLIP is susceptive to ambiguities in spatial relations and confusion of co-occurring objects, a consequence stemming from its holistic training objective.
Besides, RIM also demonstrates significant performance improvements over the most recently developed training-free methods based on visual foundation models, achieving 10.4\%, 4.2\%, and 6.6\% mIoU gains, respectively. As shown in Fig.~\ref{overall performance}, we present a series of visualizations depicting segmentation outcomes across various datasets for a more intuitive observation.

\subsection{Ablation Study}
A series of ablation studies are conducted to thoroughly investigate the impact of each component of the proposed RIM. As shown in Table~\ref{tab:ablation}, we mainly implement our experiments on COCO Object~\cite{coco} dataset, and the first row is our baseline, which follows a naive image-to-text matching for region classification. Specifically, we crop the test image along the bounding boxes corresponding to the mask proposals (generated by SAM) and resize them to $224 \times224$ resolution. The background area of the cropped image is filled with zeros, and then CLIP~\cite{clip} with ViT-B is exploited to perform image-to-text matching between the cropped image and category labels.

\textbf{Ablation study on intra-modal reference construction.} To verify the effectiveness of intra-modal alignment, we first construct a naive image-to-image matching as the $2^{nd}$ row of Table~\ref{tab:ablation}. More concretely, DINOv2 features of all synthesized images of a category are condensed to a holistic reference feature via global average pooling.
The matching process is simply implemented with cosine similarity. Despite its simplicity, this intra-modal matching exhibits a significant performance improvement, \ie, 12.0\% in mIoU over the cross-modal matching based on CLIP.
This improvement is anticipated as the holistic training objective of CLIP makes its prediction heavily reliant on contextual information. This reliance not only introduces spatial confusion but also leads to misclassification of frequently co-occurring objects, such as the $sky$ and $airplane$.
Attributed to the image generation capabilities of SD model and the robust high-level semantic feature extraction of DINOv2, an improved intra-modal alignment between region features and class reference features is achieved.

Further performance improvements can be observed if we construct the category reference features with only foreground features as shown in the $3^{th}$ row of Table~\ref{tab:ablation}.
We deem the main reason is that the background areas in synthetic images may contain information of other categories, which confuses the classification process.
While in our implementation, more refined category reference features are constructed attributed to the spatial localization ability of the SD model embedded in the cross-attention map and the powerful segmentation capability of SAM, which effectively mitigates the influence of cluttered backgrounds. Despite utilizing merely cosine similarity for intra-modal matching, it has already achieved impressive performance, \ie, 44.5\% mIoU on COCO Object dataset, which can be adopted as a strong baseline for further research.

\textbf{Ablation study on relation-aware matching.} 
As described in Sec.~\ref{sec: matching}, we propose to model the beneficial structure information contained in inter-class relationships via relation-aware matching.
Compared to the naive approach that only considers only the cosine similarity between region features and each individual category reference feature, our proposed strategy achieves a sizeable gain, \ie, 1.3\% in mIoU, as presented in the $3^{rd}$ and $4^{th}$ rows of Table~\ref{tab:ablation}.
The results prove that the proposed similarity measurement based on ranking distribution facilitates effective disambiguation in the matching process, which brings a notable reduction in the misclassification of regions.

After integrating subcategory reference features into relation-aware matching, the performance further improved from 43.1\% to 44.5\% in mIoU, as shown in $5^{th}$ row of Table~\ref{tab:ablation}. This improvement can be attributed to the ability of the SD model to generate diverse image features, enabling our method to effectively handle intra-class diversity. Moreover, the scoring scheme in Equ.~\ref{equ:10}, akin to a voting mechanism, further enhances the robustness of the matching process. Furthermore, in Table~\ref{tab:agents}, we demonstrate that reference features that share high similarities with region features are better candidates to serve as category agents.
% \begin{figure}[t]
% 	%\centering
% 	\includegraphics[width=0.98\linewidth]{images/generated_image.pdf}
% 	%\centering
% 	\vspace{-3mm}
% 	\caption{
% 		Examples of the generalized images and the corresponding masks from SAM.
% 	}
% 	\label{fig:gene}
% 	\vspace{-5mm}
% \end{figure}
\begin{figure}[t]
	%\centering
	\includegraphics[width=0.98\linewidth]{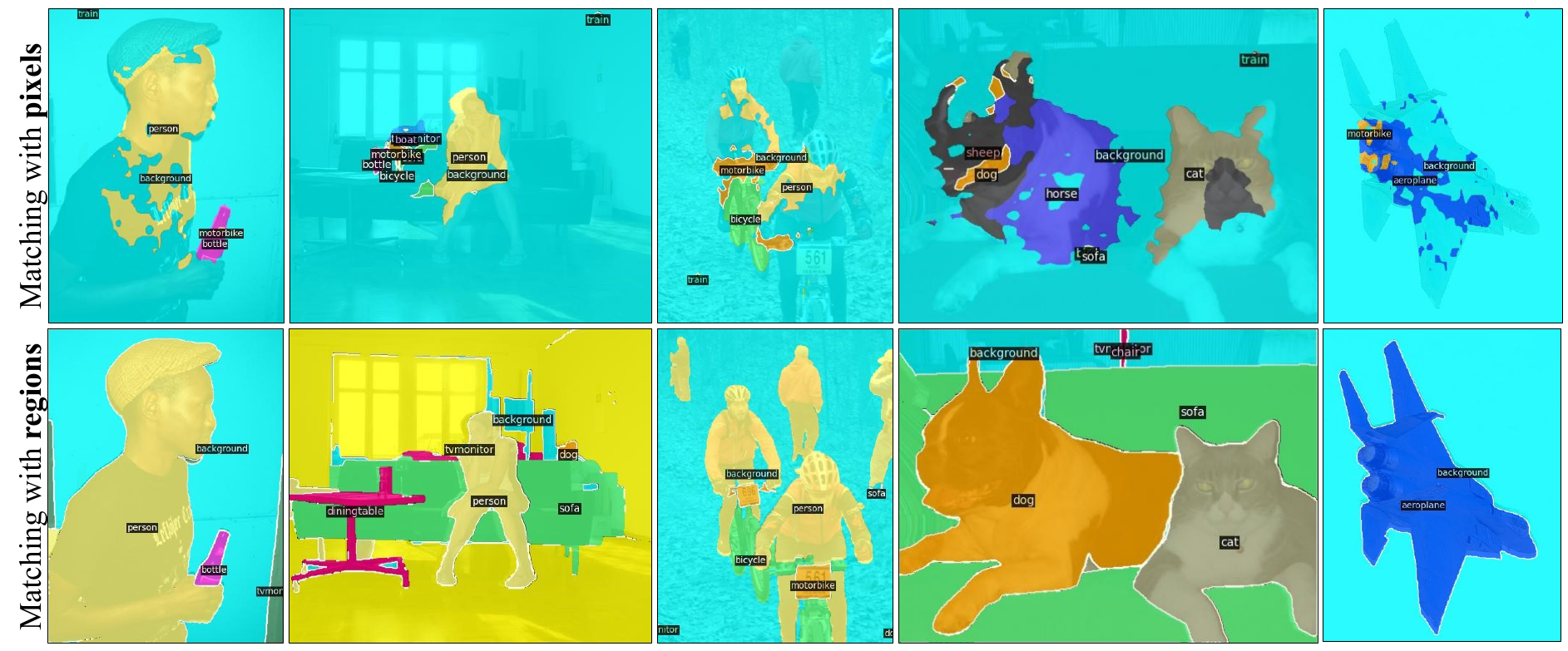}
	%\centering
	% \vspace{-3mm}
	\caption{
		Effectiveness of SAM based region-level matching. The SAM could well capture visual concepts within images.
	}
	\label{fig:prototype}
	\vspace{-5mm}
 
\end{figure}

\begin{figure}[t]
	%\centering
	\includegraphics[width=0.98\linewidth]{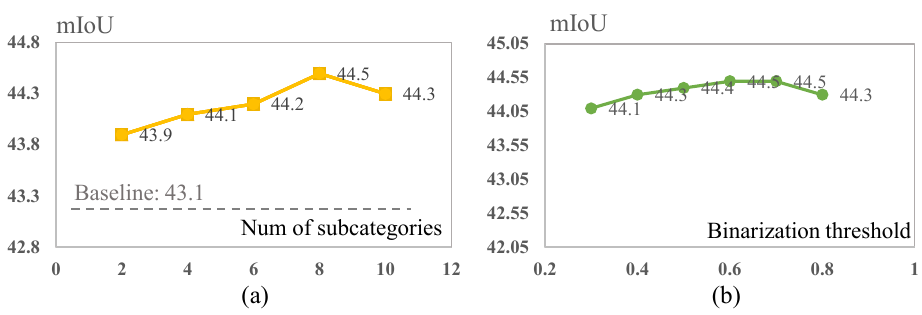}
	%\centering
	% \vspace{-3mm}
	\caption{
Hyperparameter experiments on the number of subcategories and binarization threshold of cross-attention map.
	}
	\label{fig:hyper}
	\vspace{-5mm}
\end{figure}
\textbf{Ablation on DINOv2.} 
To demonstrate the superiority of feature extraction using DINOv2, we conduct comparison experiments of CLIP~\cite{clip}, MAE~\cite{mae}, Stable Diffusion Unet (SD)~\cite{Stable_diffusion}, and DINOv2 as presented in Table~\ref{tab:encoder}.
DINOv2 achieves the best performance attribute to effective pretraining based on image-level and patch-level discriminative self-supervised learning.
While CLIP fails to match region-level features precisely, Stabel Diffusion UNet is constrained to some extent under the unconditional settings. MAE exhibits suboptimal performance in our setting due to its relatively limited high-level semantic modeling capability.

\textbf{Ablation on SAM.} 
Our relation-aware matching operates at the region level, as SAM furnishes comprehensive mask proposals for the test images.
To explore the effectiveness of SAM, on the PASCAL VOC dataset, we compare SAM with modified MaskFormer~\cite{maskformer} pre-trained on the COCO-stuff dataset following~\cite{mask-adapted-clip}. We also construct a vanilla baseline that compares category reference features with pixel features of test images directly. As shown in Table~\ref{tab:sementer}, both segmenters outperform the baseline by a large margin. We deem the main reason is that the region-level matching is more robust than pixel-level one. As illustrated in Fig.~\ref{fig:prototype}, the pixel classification paradigm fails to capture all visual concepts within images. SAM slightly surpasses MaskFormer as SAM can generate higher-quality masks on novel classes, naturally more suitable for open-vocabulary tasks.

\begin{figure}[t]
	%\centering
	\includegraphics[width=0.98\linewidth]{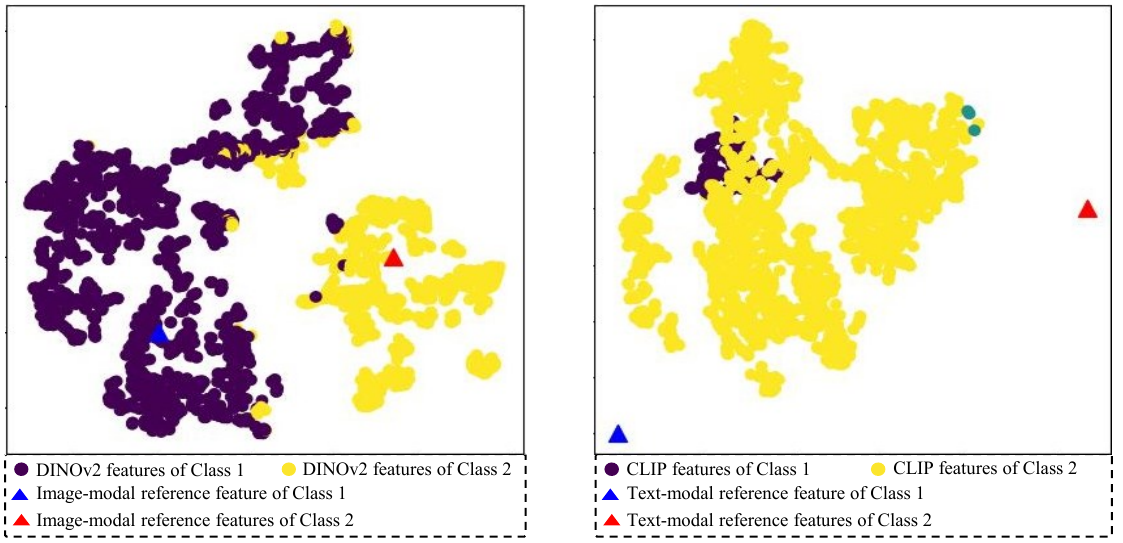}
	%\centering
	% \vspace{-3mm}
	\caption{
	T-SNE visualization of features of different region, as well as corresponding intra-modal and cross-modal reference features.}
	\label{fig:tsne}
	\vspace{-5mm}
\end{figure}

\textbf{Investigation of the intra-modal matching.}
In Fig.~\ref{fig:tsne}, we visualized the t-SNE plots of the region features alongside their corresponding text-modal reference features and image-modal reference features using the same image based on CLIP~\cite{clip} and DINOv2~\cite{oquab2023dinov2}, respectively. It can be observed that in the all-purpose feature space of DINOv2, the region features of different categories are well-differentiated and align well with their corresponding reference features. However, in the case of CLIP, different region features not only exhibit notable overlap but also fail to align effectively with corresponding text-modal reference features. This difference substantiates the underlying rationale of our motivation, \ie, intra-modal matching.

\textbf{Hyperparameter Evaluations.} Quantitative experiments are conducted to find a suitable number of subcategories used in relation-aware matching. As illustrated in Fig.~\ref{fig:hyper} (a), the performance continues to grow until the number achieves 8, beyond which it starts to decline. It is reasonable as an appropriate number of subcategory references can model category diversity, but over-partitioning may lead to deviations from instance features. We then explore how the binarization threshold affects the performance in Fig.~\ref{fig:hyper} (b). The model shows low sensitivity to threshold values owing to the capability of SAM, with a setting of 0.7 yields marginally better results.

\section{Conclusion}
\label{sec:Conclusion}
In this work, we presented a training-free Relation-aware Intra-modal Matching (RIM) framework to tackle the challenging open-vocabulary semantic segmentation. We construct better-aligned image-modal category reference features based on the Stable Diffusion model and SAM. Then a relation-aware matching strategy is employed for region classification. RIM not only achieves results significantly surpassing the state-of-the-art but also opens up new avenues for OVS from an image-to-image matching perspective.
\section{Acknowledgments}
This work was partially supported by the National Defense Basic Scientific Research Program of China (Grant JCKY2021130B016),  National Nature Science Foundation of China (Grant 12150007), Youth Innovation Promotion Association CAS 2018166.
{
    \small
    \bibliographystyle{ieeenat_fullname}
    \bibliography{main}
}

% \input{X_suppl}

\end{document}